\documentclass[conference]{IEEEtran}
\IEEEoverridecommandlockouts
\usepackage{cite}
\usepackage{amsmath,amssymb,amsfonts}
\usepackage{algorithmic}
\usepackage{graphicx}
\usepackage{textcomp}
\usepackage{xcolor}
 \usepackage{booktabs}

\def\BibTeX{{\rm B\kern-.05em{\sc i\kern-.025em b}\kern-.08em
    T\kern-.1667em\lower.7ex\hbox{E}\kern-.125emX}}
\begin{document}

\title{CFD-HAR: User-controllable Privacy through Conditional Feature Disentanglement
}


\author{\IEEEauthorblockN{Alex Gn, Fan Li, S Kuniyilh
and  Ada Axan}
\IEEEauthorblockA{}
}

\maketitle

\begin{abstract}
Modern wearable and mobile devices are equipped with inertial measurement units (IMUs). Human Activity Recognition (HAR) applications running on such devices use machine-learning-based, data-driven techniques that leverage such sensor data. However, sensor-data-driven HAR deployments face two critical challenges: protecting sensitive user information embedded in sensor data in accordance with users' privacy preferences and maintaining high recognition performance with limited labeled samples. 
This paper proposes a technique for user-controllable privacy through feature disentanglement-based representation learning at the granular level for dynamic privacy filtering.
We also compare the efficacy of our technique against few-shot HAR using autoencoder-based representation learning. We analyze their architectural designs, learning objectives, privacy guarantees, data efficiency, and suitability for edge Internet of Things (IoT) deployment. Our study shows that CFD-based HAR provides explicit, tunable privacy protection controls by separating activity and sensitive attributes in the latent space, whereas autoencoder-based few-shot HAR offers superior label efficiency and lightweight adaptability but lacks inherent privacy safeguards. We further examine the security implications of both approaches in continual IoT settings, highlighting differences in susceptibility to representation leakage and embedding-level attacks. The analysis reveals that neither paradigm alone fully satisfies the emerging requirements of next-generation IoT HAR systems. We conclude by outlining research directions toward unified frameworks that jointly optimize privacy preservation, few-shot adaptability, and robustness for trustworthy IoT intelligence.
\end{abstract}


\section{Introduction}
Data generated by wearable Internet of Things (IoT) devices are used to provide better services, such as tracking individual fitness scores, health alerts, and additional insights to users~\cite{zhang2012usc, sun2025survey, dritsas2025survey, yurtsever2020survey}. However, such data can also perform sensitive attribute inference~\cite{hu2022membership}. Research shows that sensitive user attributes, such as personal and biological data, location, and other sensitive data, can be inferred from user data shared by IoT systems~\cite{malekzadeh2019mobile, menon2025ai, almutairi2025iot}.
While user data perturbation techniques for IoT systems have been demonstrated, they have not accounted for users' dynamic privacy needs or controllability. For instance, some users may be willing to share data that can infer their gender with service providers in exchange for improved utility. Similarly, the user may prefer to keep their location private in certain contexts. 
The impact of personalization on utility in privacy-preserving IoT systems is complex.

Although personalization can enhance user experience by tailoring services to individual needs, it can also degrade performance when data required for classification tasks are restricted due to privacy concerns~\cite{jain2021differentially}. 
Such dynamic privacy requirements must be taken into account when designing a privacy-preserving technique for a data-driven IoT system. 
Perturbation has been considered a de facto standard for obfuscating sensor data to preserve user privacy, including age and gender~\cite{jain2021differentially, chathoth2022differentially}. 
Existing work aims to protect against such sensitive inference attacks by perturbing the input data by transformation or replacement, thereby degrading HAR system performance~\cite{malekzadeh2018replacement, malekzadeh2019mobile, raval2019olympus}. Moreover, these works haven't analyzed in detail the need for privacy personalization and its impact on system performance.

In general, the perturbation doesn't account for the data's detailed representations and often over-perturbs the input. In some cases, it may be sufficient to perturb only a subset of the data points. So, efficient perturbation requires careful analysis of data representation.
This raises an interesting problem: isolating the sources of variation across different data samples corresponding to the activity being performed in a HAR system by learning disentanglement with respect to utility-relevant features and the sensitive and nonsensitive attributes they generate. 

In this paper, we focus on a feature disentanglement technique that separates the underlying concepts of personal attributes from the activity being performed. Specifically, isolating the feature space corresponding to "A walking man," "A person is walking", "A tall person is running", and "A person is running, but the gender and height are not available".
Additionally, I consider combinations of such feature details, such as gender, age, and location, and isolate them from the latent space, sending only the features required for classification to the service provider for inference.
We further explore how this feature disentanglement can be used to provide better control over the level of data perturbation. Additionally, I study how such fine controls can affect model performance without compromising user-preferred privacy settings.


To manage user-controlled privacy, data-driven IoT systems can incorporate personalized privacy settings that allow users to specify which attributes they consider sensitive. These settings can then be used to adjust the level of data perturbation required while minimizing impact on utility.
In a more advanced scenario, the user may be given the option to control the privacy-utility trade-off with greater granularity by introducing additional user-specified privacy controls, such as the privacy weight each user assigns to attributes such as gender, age, and location.
Additionally, large language model (LLM)-based solutions are being deployed in IoT systems to preserve privacy by leveraging few-shot learning techniques.
While such techniques require continuous fine-tuning and training, which affects operating efficiency, they can also improve sustainability, especially when we leverage large language models in the IoT domain. Sustainability is another important design parameter to consider when deploying large language models on connected, cloud-enabled electronic devices~\cite{hussein2024ai, zhao2026electronic, mishra2025ai, hu2022membership}.

Overall, HAR in IoT and wearable environments must simultaneously address two critical, often competing requirements: protecting user privacy and operating effectively with limited labeled data. 
In this work, we present a user-controllable privacy via conditional feature disentanglement (CFD) at a granular level. We also perform a comparative analysis against another state-of-the-art technique, few-shot HAR, using autoencoder representations.
Our study proposes a CFD-based HAR that provides explicit and tunable privacy protection by structurally separating activity-relevant information from sensitive user attributes. This makes it particularly suitable for privacy-sensitive IoT applications such as healthcare monitoring and personal wearables, where regulatory compliance and user trust are paramount. 


\section{Background}

This section provides some background on techniques relevant to our paper.

\subsection{HAR Privacy}
HAR systems operate by modeling patterns in data generated by Inertial Measurement Units (IMUs) embedded in wearable devices~\cite{wang2019deep, chathoth2025privclip, chavarriaga2013opportunity}. However, the data captured by such devices can inadvertently disclose sensitive personal attributes, including age, gender, and height. Moreover, the sensor data specific to an individual can be further exploited to infer details about their private activities. For instance, data collected to recognize general walking activity may also be used to determine whether the individual is smoking while walking. These privacy risks pose significant challenges for the design and deployment of HAR systems, necessitating careful consideration of data protection and privacy-preserving modeling techniques.

\subsection{User-controllable HAR Privacy}
User-controllable privacy in HAR systems concerns constraining the system to recognize only those activities deemed acceptable by individual users, in accordance with their personal privacy preferences. These preferences may vary over time and across different contexts or locations. Consequently, dynamically adjustable, user-driven privacy controls are a critical design requirement for HAR systems. While HAR privacy has been studied in detail, less work has examined dynamic user-controllability of privacy~\cite{malekzadeh2018replacement, chathoth2025dynamic}.

\subsection{Autoencoder-based Transformation}


An autoencoder(AE) is typically used to perturb input data. Its encoder can learn to represent input features in a low-dimensional latent space, which the decoder can then use to reconstruct the input.
An autoencoder learns to reconstruct the input based on the reconstruction loss.
A replacement autoencoder is a transformation method that first learns a mapping from sensitive to nonsensitive data, and then replaces discriminative features corresponding to sensitive inferences with features more commonly observed in nonsensitive inferences. This approach only works when the sensitive and nonsensitive data are clearly separated. Moreover, this method does not consider utility~\cite{malekzadeh2018replacement}.
Olympus is a utility-aware obfuscation method that models utility and privacy requirements as adversarial networks, thereby hiding private information in user data with minimal utility loss~\cite {raval2019olympus}.
However, these privacy benefits come at the cost of increased training complexity, reliance on sensitive attribute annotations, and potentially higher computational overhead at the edge.

In contrast, AE-based few-shot HAR prioritizes data efficiency and lightweight deployment~\cite{10.1007/978-981-96-2468-3_16}. Leveraging unsupervised representation learning enables rapid adaptation in low-label regimes common in IoT environments. This makes AE-based approaches attractive for resource-constrained edge devices and personalized activity recognition scenarios. Nevertheless, the absence of explicit privacy constraints leads to entangled latent representations that may inadvertently encode sensitive user information, increasing the risk of privacy leakage and representation-level attacks.

While the above methods provide privacy by obfuscating private data, data transformation can degrade utility. We propose a different approach to privacy preservation by considering fine-grained user controls over privacy requirements and conditioning the transformation function on the disentanglement representation corresponding to those controls.

\subsection{Few-Shot HAR Using Autoencoder Representations}

Few-shot HAR approaches focus on improving recognition performance under scarce labeled data. Autoencoder models are commonly used to learn compact latent representations from large volumes of unlabeled sensor data \cite{hinton2006reducing}. The encoder $f_\theta$ maps input $x$ to latent vector $z$, while the decoder reconstructs the signal:

\begin{equation}
\mathcal{L}_{AE} = \|x - \hat{x}\|_2^2.
\end{equation}

The learned embedding is then used for few-shot classification using metric learning or lightweight classifiers.

AE-based HAR is attractive for IoT because it is label-efficient, computationally lightweight, and well-suited to on-device learning. It enables rapid personalization and adaptation in environments where annotated data is scarce or expensive to obtain.
Nevertheless, the autoencoder objective preserves all informative factors in the signal, including sensitive attributes. 
Consequently, AE-based HAR representations may encode strong user fingerprints, making them vulnerable to privacy attacks, including membership and attribute inference.

\section{Design}

\subsection{Threat Model}
\begin{figure}[t]
    \centering
    \includegraphics[width=3.5in]{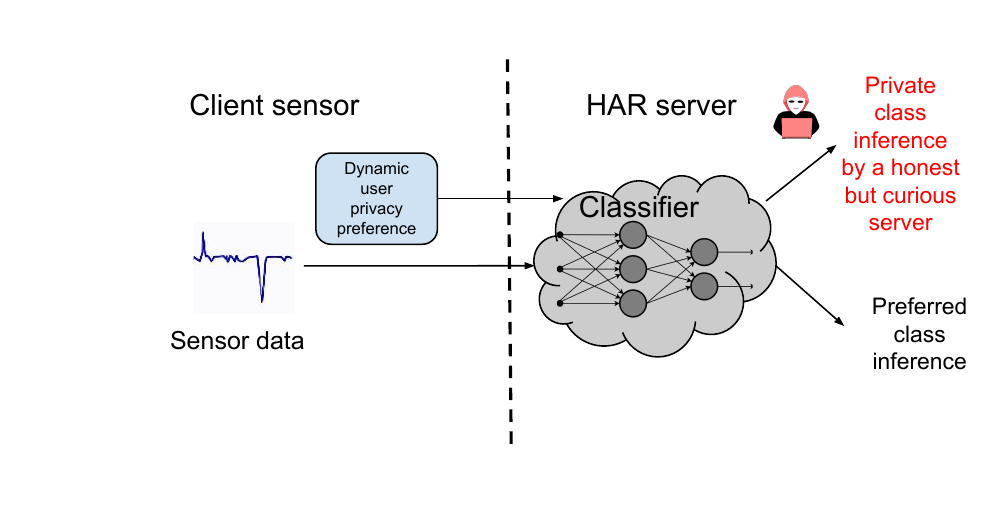} 
    \caption{Threat model}
    \label{fig:CFD-HAR-TM}
\end{figure}

We consider a setting in which client devices continuously generate sensor data, which are subsequently transmitted to a centralized server hosting an ML–based HAR model. In our threat model, the wearable devices are assumed to be honest, whereas the centralized server is modeled as honest-but-curious, i.e., it correctly follows the prescribed protocol but may attempt to infer additional information from the received data without altering it. We depict our threat model in Figure~\ref{fig:CFD-HAR-TM}.

\subsection{Problem statement}
A dataset $D$ contains motion sensor readings from different users, and a model $M$ is trained using the dataset $D$.
Let $S$ be a set of user attributes, such as age, gender, height, weight, location, and personal habits. $S_u$ is a set of user-sensitive characteristics of a user $u$. The goal is to perturb the data $D_u$ of user $u$ such that an attacker $A$ who intercepts the model inference process can not predict the sensitive attributes $S_u$ of user $u$ from the data $D_u$, while inferring the attributes from $S$ that are not in $S_u$ is acceptable.
Let $X_u$ be the latent representation of the activities performed by a user $u$ corresponding to the row sensor data $D_u$. Apart from the representation $G_u$ specific to the activity being performed, the latent representation may also include other redundant representations, including representations corresponding to user-sensitive attribute $H_u$ tangled to it. Thus, the overall latent representation can be written as $X_u = G_u + H_u$, and the $H_u$ component, if left untangled, enables the attacker to perform sensitive-attribute inference.
So the goal is to remove or disentangle the latent representation of user-sensitive attributes $H_u(W_u(i))$ from $X_u$ according to the privacy weights $W_u(i)$ associated with each attribute $i$. 

\subsection{Feature Disentanglement}
Unlike prior work on privacy preservation, we adopt a different approach by providing dynamic controls through feature disentanglement.
Client-level feature disentanglement can filter out local nodes' privacy attributes in federated settings~\cite{zhou2023privacy}.
A variational autoencoder(VAE) represents the features in a latent representation and forces the latent space to be as close as the prior by minimizing the KL divergence between the posterior and the prior as much as possible~\cite{van2014renyi}. This maximizes the probability of generating real data while keeping the distance between the real and approximate posterior distributions small, which amounts to keeping the distance between the posterior and prior small~\cite{higgins2017beta}.
However, a VAE can't guarantee that the selected point from the latent space is an input similar to what we are looking for. For instance, in a handwritten digit generation system trained on digits from 0 to 9, we can't ask the VAE to select and reconstruct a particular digit, say 1.
Additionally, there is a trade-off between disentanglement and the VAE's reconstruction capability~\cite {chen2016infogan}.
$\beta$-VAE is a type of variational autoencoder that seeks to discover disentangled latent factors. It modifies VAEs by introducing an adjustable hyperparameter that balances latent-channel capacity and independence constraints against reconstruction accuracy~\cite{higgins2017beta}.

\begin{figure*}[t]
    \centering
    \includegraphics[width=6in]{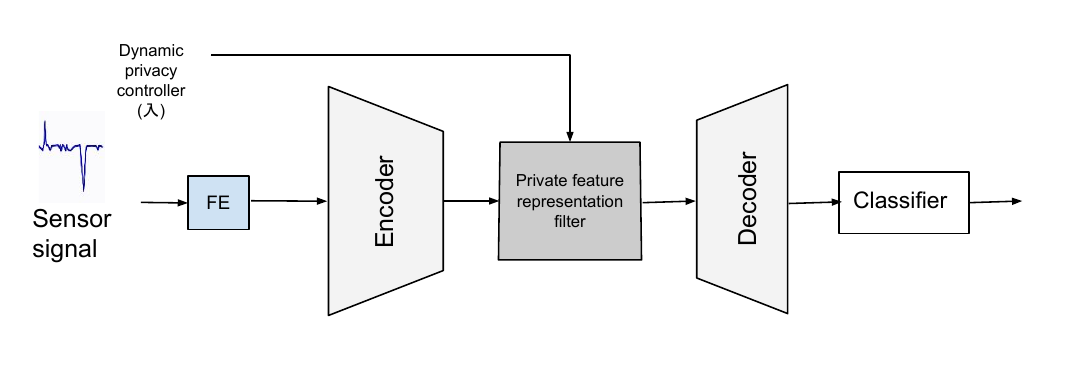} 
    \caption{CFD-HAR Architecture.}
    \label{fig:CFD-HAR-Archi}
\end{figure*}

Conditional VAE (CVAE) can generate samples from the learned latent space conditioned on given inputs, providing greater user control over the generated data. The encoder processes the input data along with its conditions, while the decoder uses the resulting conditional latent representation to reconstruct the data or generate new instances conditioned on specific attributes.
Methods have been demonstrated to extract semantic-rich temporal correlations from the latent representations of time-series data by leveraging disentanglement techniques~\cite {li2022towards}.
While researchers have explored learning disentangled behavior patterns for wearable-based human activities~\cite{zhou2023conditional}, no work has addressed providing user-controllable data privacy based on disentanglement representations of human activities~\cite{su2022learning}.



\section{Privacy-Preserving HAR via Conditional Feature Disentanglement}

The architecture of privacy-preserving Conditional feature Disentanglement HAR (CFD-HAR) is depicted in Figure ~\ref{fig:CFD-HAR-Archi}.
Conditional feature disentanglement aims to explicitly separate task-relevant activity information from sensitive user attributes. Given an input sensor sequence $x$, the encoder learns a structured representation:

\begin{equation}
z = [z_{\text{activity}}, z_{\text{privacy}}],
\end{equation}

where $z_{\text{activity}}$ captures activity semantics and $z_{\text{privacy}}$ captures sensitive factors. Training involves a multi-objective loss function:

\begin{equation}
\mathcal{L} =
\mathcal{L}_{\text{activity}}
+ \lambda \mathcal{L}_{\text{privacy}},
\end{equation}

where the privacy term often employs adversarial learning to minimize leakage of sensitive attributes \cite{edwards2015censoring}. This framework enables user-controllable privacy by adjusting the trade-off parameter $\lambda$.

We provide fine control on the $\lambda$ by considering privacy importance to each attribute $j$ and for each activity $i$ as shown in the equation below.

\begin{equation}
\mathcal{L} =
\Sigma \mathcal{L}_{\text{activity}_{i}}
+ {\lambda}_{j} \mathcal{L}_{\text{privacy}_{j}},
\end{equation}

We use the condition parameter ${\lambda}_{1} {\lambda}_{2} ... {\lambda}_{j}... {\lambda}_{n} $ as an input for dynamically controlling the user's privacy preference, which can be entered as a condition input for a CVAE-based autoencoder. We use the CVAE model explained in the previous section.

In IoT HAR settings, CFD offers several advantages. First, it directly addresses privacy leakage, a critical concern for wearable sensing applications subject to regulatory requirements. Second, the structured latent space improves interpretability and downstream transfer. Third, adversarial disentanglement can reduce identity-specific overfitting.

However, CFD typically requires access to sensitive attribute labels during training and may suffer from imperfect disentanglement, especially under limited data or strong distribution shift. Moreover, the adversarial training process increases computational overhead, which may be challenging for resource-constrained edge devices.

\begin{table*}[t]
\centering
\caption{Comparison Between Conditional Feature Disentanglement HAR and Few-Shot Autoencoder-Based HAR}
\label{tab:cfd_vs_fewshot_har}
\renewcommand{\arraystretch}{1.15}
\begin{tabular}{|p{3.2cm} |p{5.3cm} |p{5.3cm}|}
\hline
\textbf{Aspect} & \textbf{CFD-HAR} & \textbf{FS-HAR} \\
\hline

Primary Objective &
User-controllable privacy  &
Few-shot learning \\\hline

Learning Paradigm &
conditional disentanglement &
Few-shot  \\\hline

Latent Representation &
Disentanglement &
Compact but entangled \\\hline

Label Requirements &
Requires activity labels and often sensitive attribute labels &
Requires few activity labels; no sensitive labels needed \\\hline

Privacy Protection &
\textbf{Strong and user-controllable} &
Limited or absent \\\hline

Sustainability &
Moderate computation&
High computation\\

\hline
\end{tabular}
\end{table*}

Compared with the few-shot HAR, the CFD-HAR approach reflects different design priorities from an IoT deployment perspective. CFD-based HAR prioritizes privacy preservation by enforcing structured latent separation, whereas AE-based few-shot HAR prioritizes data efficiency and lightweight adaptation. In edge scenarios with strict privacy requirements—such as healthcare wearables—CFD provides stronger protection against the leakage of sensitive information. In contrast, in low-resource environments with extremely limited labeled data, AE-based methods may offer superior recognition performance with minimal supervision.

However, the two paradigms also differ in their security posture. Because CFD explicitly constrains representation structure, it may partially mitigate certain forms of representation leakage. In contrast, the unconstrained latent space of autoencoders is more susceptible to embedding manipulation and backdoor insertion. In continual IoT settings, replay and incremental updates may further amplify these risks.

We provide a comparative analysis in Table~\ref{tab:cfd_vs_fewshot_har}. In summary, conditional feature disentanglement and autoencoder-based few-shot HAR represent complementary approaches that address privacy and data-efficiency challenges, respectively. Their differing objectives lead to distinct trade-offs in privacy protection, computational cost, sustainability, and security robustness. Understanding these trade-offs is essential for designing trustworthy HAR systems in emerging IoT environments.

\section{Experimental setup}
In this section, we describe the experimental setup used.

\subsection{Dataset}
We use two datasets of human activities.
\begin{enumerate}
\item Motion-sense dataset consists of 6 activities and 4 personal attributes ~\cite{malekzadeh2018replacement}.

\item
Daily and Sports Activities Dataset(DSADS)  comprising 19 activities with location attribute~\cite{barshan2014recognizing}.
\end{enumerate}

\subsection{Model}
We use a feature extractor and CVAE with an autoencoder, a feature filter, and a decoder.
We consider 4 attributes to be private according to user preferences. The output of CVAE is sent to two classifiers. One is an activity classifier, and the other is an attribute classifier.
The encoder layer has 3 neural networks and a latent representation dimension of 2. The decoder mirrors the encoder, taking an additional input to select the privacy label and choose the appropriate latent representation for generating filtered sensor data as output. KL divergence is used as a loss function. We use the classifiers the same as those used by the Olympus paper~\cite{raval2019olympus}.

\section{Evaluation}

We conduct our experiments by varying combinations of private attributes and measure activity and identification performance. Identity recognition is based on the predicted private attributes of each individual.
Below, we discuss the evaluation of each experiment in detail.

\subsection{Single private attribute}

\begin{figure}[t]
    \centering
    \includegraphics[width=3.5in]{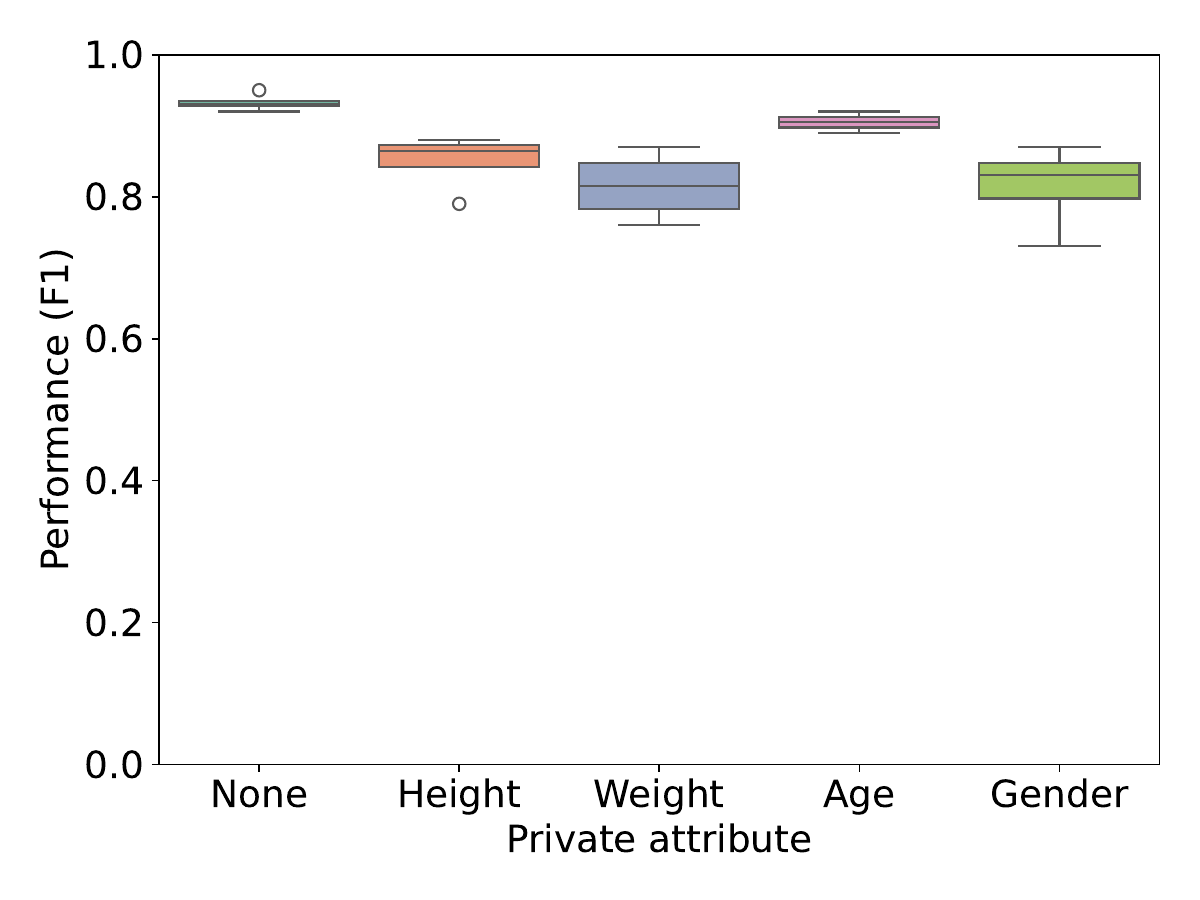} 
    \caption{Activity classification performance against each private attribute preference.}
    \label{fig:CFD-HAR-F1-utility}
\end{figure}
\begin{figure}[t]
    \centering
    \includegraphics[width=3.5in]{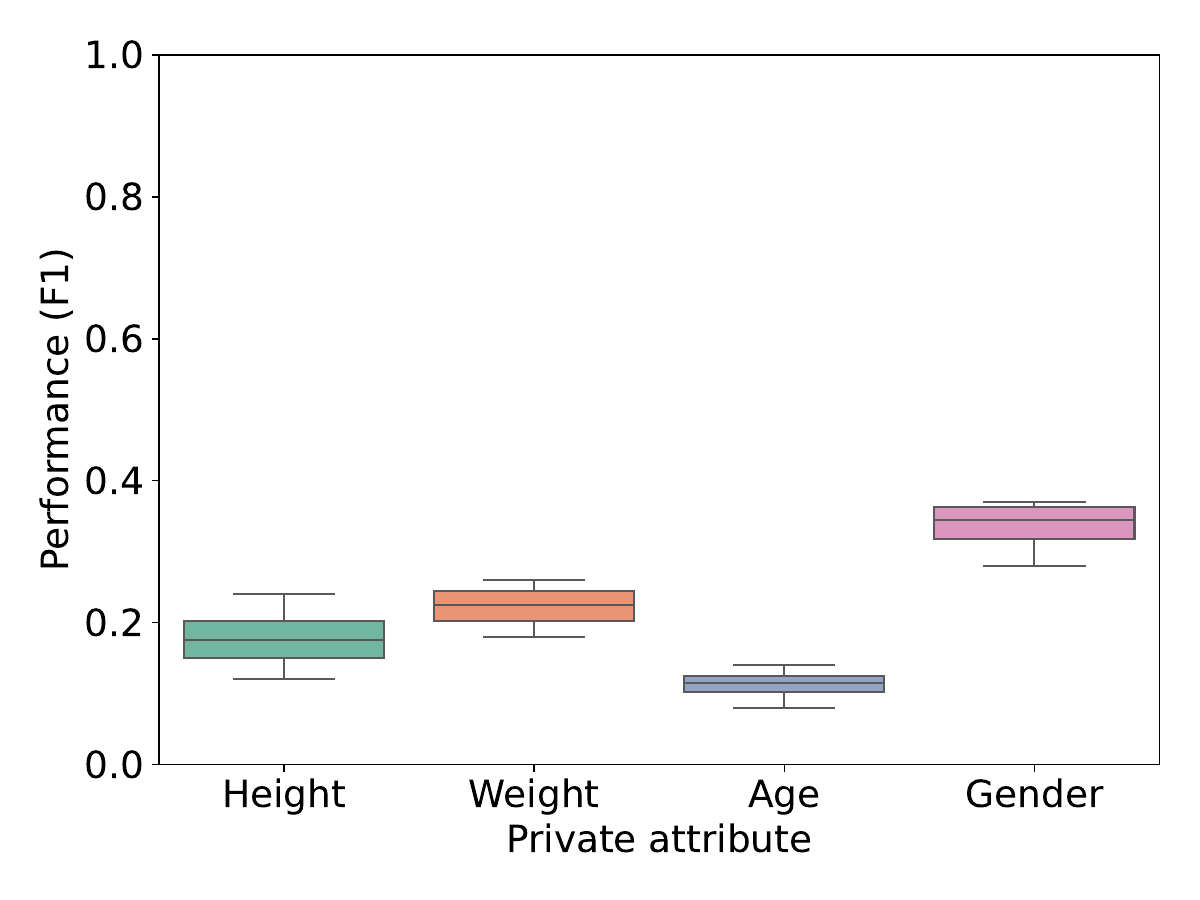} 
    \caption{Re-identification performance against each private attribute preference.}
    \label{fig:CFD-HAR-F1-privacy}
\end{figure}

We first consider one attribute at a time as private and run our experiments, and the results of classification performance and user re-identification performance are plotted in Figure~\ref{fig:CFD-HAR-F1-utility} and ~\ref{fig:CFD-HAR-F1-privacy} respectively. As shown, the model achieves high performance while maintaining attribute-based identification, as evidenced by the low F1-score in Figure~\ref{fig:CFD-HAR-F1-privacy}.

\subsection{Multiple private attributes}
Next, we consider dynamically selecting combinations of private attributes and running the model to infer activity and identity from the attribute predictions. We tabulate the results in Table~\ref{tab:Activiyidentity}. We code each attribute combination as 4 bits, with each bit representing 'height', 'weight', 'age', and 'gender' in that order. For instance, '0101' indicates that the user prefers 'weight' and 'gender' to be private, whereas 'height' and 'age' are non-sensitive.

\begin{table*}[t]
    \centering
    \caption{Activity classification performance in F1-score based on various private activity combinations selected dynamically.}    
    \begin{tabular}{|l|c|c|c|c|c|c|c|c|l|l|l|l|l|l|l|l|}\hline
                 &0000 & 0001 &  0010&  0011&  0100&  0101&  0110&  0111&     1000 &1001 &1010& 1011& 1100& 1101& 1110&1111\\\hline
                  Activity& 0.94&   0.84& 0.88 & 0.83 & 0.86 &  0.82 & 0.87 &  0.79&0.93 & 0.85 &0.89& 0.85& 0.84& 0.82 & 0.85&0.77\\\hline
                  Identity& \textbf{0.79} & 0.24 & 0.19 & 0.18 & 0.21 & 0.31 & 0.22 & 0.23 &  0.33   & 0.27& 0.17& 0.19&0.25 & 0.19& 0.13&\textbf{0.11}\\\hline

    \end{tabular}

    \label{tab:Activiyidentity}
\end{table*}

\subsection{Varying importance of privacy requirements}
\begin{figure}[t]
    \centering
    \includegraphics[width=3.5in]{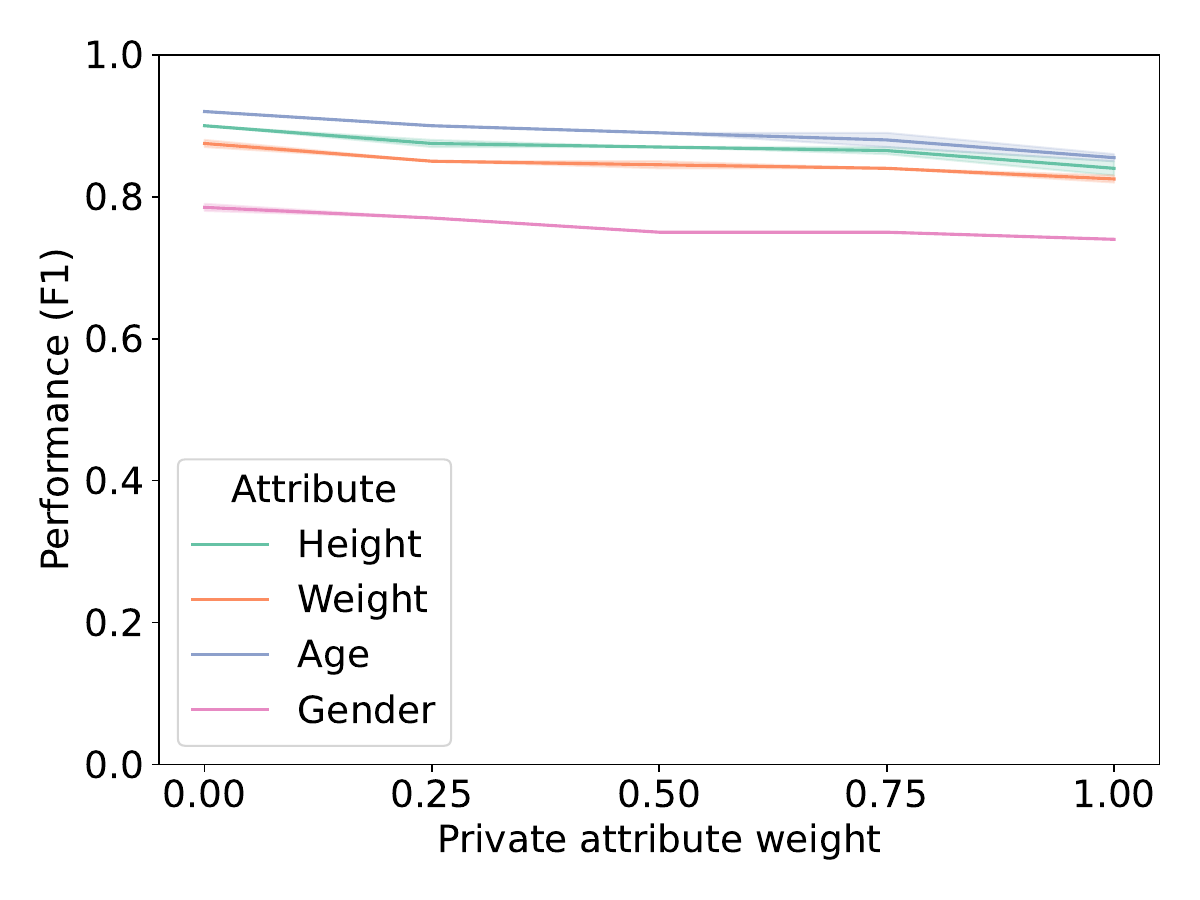} 
    \caption{Activity classification performance against each private attribute weight.}
    \label{fig:CFD-HAR-F1-utility-line}
\end{figure}

\begin{figure}[ht]
    \centering
    \includegraphics[width=3.5in]{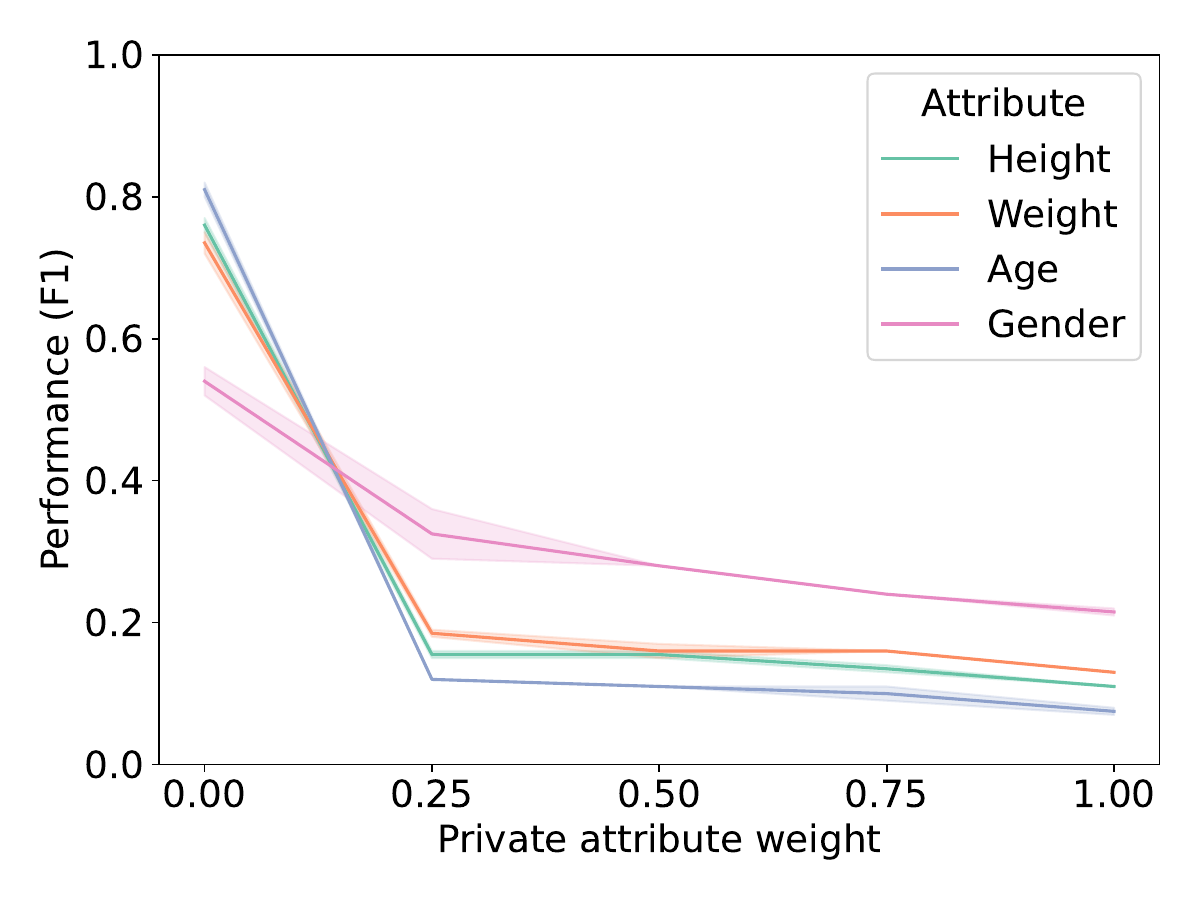} 
    \caption{Re-identification performance against each private attribute weight.}
    \label{fig:CFD-HAR-F1-privacy-line}
\end{figure}

We vary the weight of each attribute-related privacy preference from 0 to 1, where 0 indicates low sensitivity and 1 indicates high sensitivity for the user. 
Figures~\ref{fig:CFD-HAR-F1-utility-line} and \ref{fig:CFD-HAR-F1-privacy-line} show the impact of performance in terms of activity recognition and identification based on attributes, as privacy weight changes for each attribute, taking one at a time.
The results show that as the weight of the privacy attribute increases, the reidentification becomes more difficult without degrading the activity classification to the same extent.

\section{Future Directions}

For IoT systems employing continual or contrastive continual learning, the choice between CFD and AE-based few-shot learning has important security implications. CFD can reduce the coupling between sensitive attributes and activity representations, potentially limiting certain privacy attacks. However, imperfect disentanglement may still allow adversarial triggers to persist in the task subspace. AE-based few-shot HAR, while efficient, presents a larger attack surface due to its dense and unconstrained latent representations.

A promising future direction is to integrate few-shot adaptation with privacy-aware disentanglement and contrastive objectives. Such hybrid frameworks could simultaneously address label scarcity, preserve privacy, and be robust to representation-level attacks—an increasingly important requirement for next-generation IoT HAR systems.


\section{Related work}

IoT devices play a major role in enabling smartness and intelligence by automating processes in embedded systems, which are widely deployed in health monitoring and home automation~\cite{chavarriaga2013opportunity, melnyk2025hardware, zhang2012usc, wanyan2024comprehensive, alfahaid2025machine}.
Modern connected applications generate large volumes of data from sensor events, audit logs, data-traffic logs, error logs, alarms, and network-traffic logs~\cite{le2021log, peng2025log, javanmardi2025integration, dritsas2025survey}.
While data analytics is important for cloud-based services, leveraging AI and ML tools to enhance application quality, security, and privacy remains a serious concern. While ML models can be backdoored, the data used for training such models can also poses threat when such data is used for
user-sensitive attribute inference in data-driven systems.
Several techniques, such as differential privacy and homomorphic encryption, have been proposed, but each has limitations. While DP claims to provide a privacy guarantee at the expense of utility, more granular user controls are often ignored in DP-based proposed solutions. DP perturbs the data shared by the user, but our goal is to share only the data required for the utility while preventing the attacker from inferring sensitive attributes~\cite{chathoth2021federated}.

An autoencoder is typically used to perturb input data for privacy preservation or data poisoning in adversarial learning~\cite{chathoth2024dynamic}. Its encoder learns to represent input features in a low-dimensional latent space, which the decoder then uses to reconstruct the input. The reconstruction capability of an autoencoder can also be used for anomaly detection by evaluating the reconstruction loss and measuring deviations from the statistics of benign data~\cite {chen2018autoencoder, chathoth2025pcap}.
An autoencoder learns to reconstruct the input using a reconstruction loss.
A replacement autoencoder is a transformation method that first learns a mapping from sensitive to nonsensitive data, and then replaces discriminative features corresponding to sensitive inferences with features more commonly observed in nonsensitive inferences. This approach only works when the sensitive and nonsensitive data are clearly separated. Moreover, this method does not consider utility~\cite{malekzadeh2018replacement}.
Olympus is a utility-aware obfuscation method that models utility and privacy requirements as adversarial networks, thereby hiding private information in user data while minimizing utility loss~\cite {raval2019olympus}.
Few-shot learning techniques are used to preserve privacy because they do not require sensitive data for training. However, the performance of such models based on autoencoders or contrastive learning is not particularly strong~\cite{feng2019few, ganesha2024few,chathoth2026contrastive, wanyan2024comprehensive, ruan2024advances}. 

While the above methods provide privacy by obfuscating sensitive data, data transformation can degrade utility~\cite{malekzadeh2020privacy, chathoth2025privclip}. We propose a different approach that provides a privacy-preserving technique by considering fine-grained user controls over privacy requirements and conditioning the transformation function on the disentanglement representation corresponding to those controls.
Client-level feature disentanglement can filter out local nodes' privacy attributes in federated settings~\cite{zhou2023privacy}.
A variational autoencoder(VAE) represents the features in a latent representation and forces the latent space to be as close as the prior by minimizing the KL divergence between the posterior and the prior as much as possible~\cite{van2014renyi}. This maximizes the probability of generating real data while keeping the distance between the real and approximate posterior distributions small, which in turn depends on keeping the distance between the posterior and prior small.
However, a VAE can't guarantee that the selected point from the latent space is an input similar to what we are looking for. For instance, we can't ask VAE to select digit 1 in digit generation.
Additionally, there is a trade-off between disentanglement and the VAE's reconstruction capability~\cite {chen2016infogan}.
$\beta$-VAE is a type of variational autoencoder that seeks to discover disentangled latent factors. It modifies VAEs by introducing an adjustable hyperparameter that balances latent-channel capacity and independence constraints against reconstruction accuracy~\cite{higgins2017beta}.

CVAE is a conditional variational autoencoder that can generate samples from the learned latent space conditioned on given inputs, thereby providing greater user control over the generated data. The encoder processes the input data along with its conditions, while the decoder uses the resulting conditional latent representation to reconstruct the data or generate new instances conditioned on specific attributes.
Methods have been shown to extract semantically rich temporal correlations from the latent representations of time-series data by leveraging disentanglement techniques~\cite{li2022towards}.
While researchers have explored learning disentangled behavior patterns for wearable-based human activities~\cite{zhou2023conditional}, no work has addressed user-controllable data privacy based on disentanglement representations of human activities~\cite{su2022learning}.


\section{Conclusion}
\vspace{12pt}

In this paper, we present a user-controllable privacy via conditional feature disentanglement (CFD) at a granular level. We also perform a comparative analysis against another state-of-the-art technique, few-shot HAR, using autoencoder representations.
From a security standpoint, the two paradigms present different risk profiles. This work proves that CFD can help mitigate the risk of identity leakage when disentanglement is implemented effectively. On the other hand, autoencoder-based few-shot Human Activity Recognition approaches have a larger attack surface due to the dense, unconstrained embeddings. 
Overall, neither approach alone fully meets the emerging requirements of next-generation IoT HAR systems. Future research should focus on developing unified frameworks that simultaneously optimize for privacy preservation, few-shot adaptability, and robustness against attacks targeting representations. Combining contrastive learning, adversarial disentanglement, and continual adaptation appears to be a promising direction toward achieving trustworthy and efficient HAR in dynamic IoT environments.

\bibliographystyle{IEEEtran}
\bibliography{bib}

\end{document}